\pdfoutput=1

\documentclass[11pt]{article}

\usepackage{EMNLP2022}

\usepackage{times}
\usepackage{latexsym}

\usepackage{microtype}
\usepackage{inconsolata}
\usepackage{graphicx}
\usepackage[utf8]{inputenc} 
\usepackage[T1]{fontenc}    
\usepackage{hyperref}       
\usepackage{url}            
\usepackage{booktabs}       
\usepackage{amsfonts}       
\usepackage{nicefrac}       
\usepackage{microtype}      
\usepackage{algorithm}
\usepackage{algpseudocode}
\usepackage[hang]{footmisc}
\usepackage{subfigure}

\usepackage{xspace}
\usepackage{multirow}
\usepackage{rotating}

\usepackage{amsthm}
\usepackage{amsmath,amssymb,enumerate}

\usepackage{thmtools}
\usepackage{thm-restate}

\newcommand{\todoc}[2][]{\todo[color=Apricot!10,#1]{#2}}
\newcommand{\todoar}[2][]{\todo[color=Blue!30,#1]{#2}}
\renewcommand{\todoc}[2][]{}
\renewcommand{\todoar}[2][]{}

\usepackage[disable,backgroundcolor = White,textwidth=\marginparwidth]{todonotes}

\def\inl{{\langle}}
\def\inr{{\rangle}}

\def\inl{{\langle}}
\def\inr{{\rangle}}

\def\beqa{\begin{eqnarray}}
\def\eeqa{\end{eqnarray}}
\def\beqann{\begin{eqnarray*}}
\def\eeqann{\end{eqnarray*}}

\newcommand{\loss}{\ell}

\def\inl{{\langle}}
\def\inr{{\rangle}}

\renewcommand{\epsilon}{\varepsilon}

\title{BERT for Long Documents: A Case Study of Automated ICD Coding}

\author{Arash Afkanpour \And
        Shabir Adeel\thanks{Equal contribution} \And
        Hansenclever Bassani\footnotemark[1] \AND
        Arkady Epshteyn\footnotemark[1] \And
        Hongbo Fan\footnotemark[1] \And
        Isaac Jones\footnotemark[1] \And
        Mahan Malihi\footnotemark[1] \AND
        Adrian Nauth\footnotemark[1] \And
        Raj Sinha\footnotemark[1] \And
        Sanjana Woonna\footnotemark[1] \And
        Shiva Zamani\footnotemark[1] \AND
        Elli Kanal\thanks{Technical leadership} \And
        Mikhail Fomitchev\footnotemark[2] \And
        Donny Cheung\footnotemark[2]  \AND
  Google \\
  \texttt{arashaf@google.com}
}

\begin{document}
\maketitle
\begin{abstract}
Transformer models have achieved great success across many NLP problems. However, previous studies in automated ICD coding concluded that these models fail to outperform some of the earlier solutions such as CNN-based models. In this paper we challenge this conclusion. We present a simple and scalable method to process long text with the existing transformer models such as BERT. We show that this method significantly improves the previous results reported for transformer models in ICD coding, and is able to outperform one of the prominent CNN-based methods.
\end{abstract}

\section{Introduction}
\label{intorduction}

The International Classification of Diseases (ICD) codes provide a standard way of keeping track of diagnoses and procedures during a patient visit. These codes are used worldwide for epidemiological studies, billing and reimbursement, and research in health care. The codes are maintained by the World Health Organization (WHO) and are revised and updated periodically. As of 2022 the ICD codes are in the 11th revision.

Assigning ICD codes to a clinical note, such as a discharge summary, is done by professional medical coders. Human coders require extensive training, and the process of coding is often time-consuming, costly, and error-prone. Due to these challenges there is an incentive to automate the coding process. Therefore in recent years this problem has gained interest among machine learning researchers in health care (See, \citet{mullenbach2018explainable, li2020icd, zhang2020bert} and references therein). On the surface, the problem can be considered as a multi-label document classification problem. However, there are aspects of the problem that make it particularly challenging. The primary challenge is that there are tens of thousands of classes. For instance, billable ICD-10-CM codes consist of approximately 73,000 codes. In addition, the distribution of the codes is not uniform. Many of the codes are related to rare conditions and are mentioned infrequently in text, which makes it difficult to train a reliable classifier for them.

Transformer-based language models developed based on self attention \citep{vaswani2017attention} have become the state-of-the-art across many NLP problems by outperforming previous solutions that were mostly based on recurrent neural networks (RNN) and convolutional neural networks (CNN). So one would expect that they perform well in ICD coding too. However, examining the literature of ICD coding methods reveals that transformer-based solutions fail to outperform CNN-based models. Many studies have applied the BERT language model \citep{devlin2018bert} to this task, for example \citet{pascual2021towards, singh2020multi, biseda2020prediction, amin2019mlt}. More recently, \citet{ji2021does} performed a comprehensive quantitative study to compare BERT and some of its variants pre-trained on medical text against CNN-based models such as \citet{mullenbach2018explainable} and \citet{cao2020hypercore} to answer the question of whether the magic of BERT (as observed across many NLP problems) also applies to automated ICD coding. They concluded that BERT cannot outperform CNN-based models in the full ICD code case.

Unlike RNN or CNN models, which in theory can process sequences of arbitrary length, transformers' computational complexity scales quadratically with sequence length. This means that most of these models can handle limited size sequences. For instance, BERT models usually are pre-trained and fine-tuned on sequences with at most 512 tokens. Clinical notes normally contain long snippets of text beyond the sequence limit of transformers. We hypothesize that this constraint could explain the poor performance of transformers in this task, and will present empirical evidence for that.

We emphasize that we do not claim to achieve state-of-the-art performance in ICD coding, or that our design is the most efficient transformer architecture for processing long text. For a review of efficient transformers see \citet{tay2020efficient} and references therein. Our goal is to provide new empirical evidence that shows even the standard transformer models can outperform some of the previous prominent methods and are a viable solution for ICD coding.

\section{Related work}
\label{related_work}

\citet{medori2010machine} applied a rule-based method to extract important snippets of text and encode them with ICD codes. \citet{perotte2014diagnosis} proposed SVM classification with bag-of-words features. They experimented with both flat SVM (i.e. one classifier per code) and a hierarchical classifier.

With the success of deep learning in NLP tasks, many researchers focused on using RNN and CNN models for ICD coding.  CNN models provide a convenient way to learn a contextual representation of text in NLP problems \citep{chen2015convolutional}. For example, \citet{mullenbach2018explainable} proposed the \textit{CAML} model: a convolutional layer on word2vec embedding vectors to learn a contextual representation for each word. The word representations are combined into a class-specific document representation using the attention mechanism. They also suggested a method to leverage code descriptions via a regularization term. \citet{li2020icd} proposed Multi-filter Residual CNN (MultiResCNN) that uses convolutional layers with different kernel sizes to capture patterns with different lengths. Additionally, they used residual blocks on top of the convolutional layer. Similar to \citet{mullenbach2018explainable} they employed a per-class attention mechanism to make the document representation attend to different parts of the input for each code.

Recurrent neural networks (RNN) are also studied extensively for ICD coding. \citet{shi2017towards} applied LSTM at character and word level to encode both the clinical note and the code description. \citet{baumel2018multi} employs a two-layer bidirectional Gated Recurrent Unit (GRU) model, where the first layer encodes individual sentences, and the second layer encodes the document. 

With the success of transformer architectures across many NLP tasks, researchers focused their attention to designing such models for ICD coding. BERT-XML \citep{zhang2020bert} with access to a large corpus of private data managed to pre-train the model with sequence length of 1024. Most of the work in this area, however, considered the standard BERT model and its variants pre-trained on medical text to encode the document \citep{pascual2021towards, singh2020multi, biseda2020prediction, amin2019mlt}. One observation with these models was that they were unable to outperform CNN-based models. \citet{ji2021does} performed a comprehensive study to answer a few research questions on the suitability of BERT models for ICD coding. They studied and compared different variants of BERT pre-training. They also proposed a hierarchical attention method so that long clinical notes can be processed with a BERT model with a limit of 512 tokens. Most importantly, they compared different BERT variants against traditional CNN-based models, and through extensive experiments showed that BERT-based models are not capable of outperforming CNN-based models in ICD coding. In the next sections we show that a simple method that enables processing of long text with transformers will attain results that contradict the findings of \citet{ji2021does}.

\begin{figure}[t]
    \centering
    \includegraphics[width=0.45\textwidth]{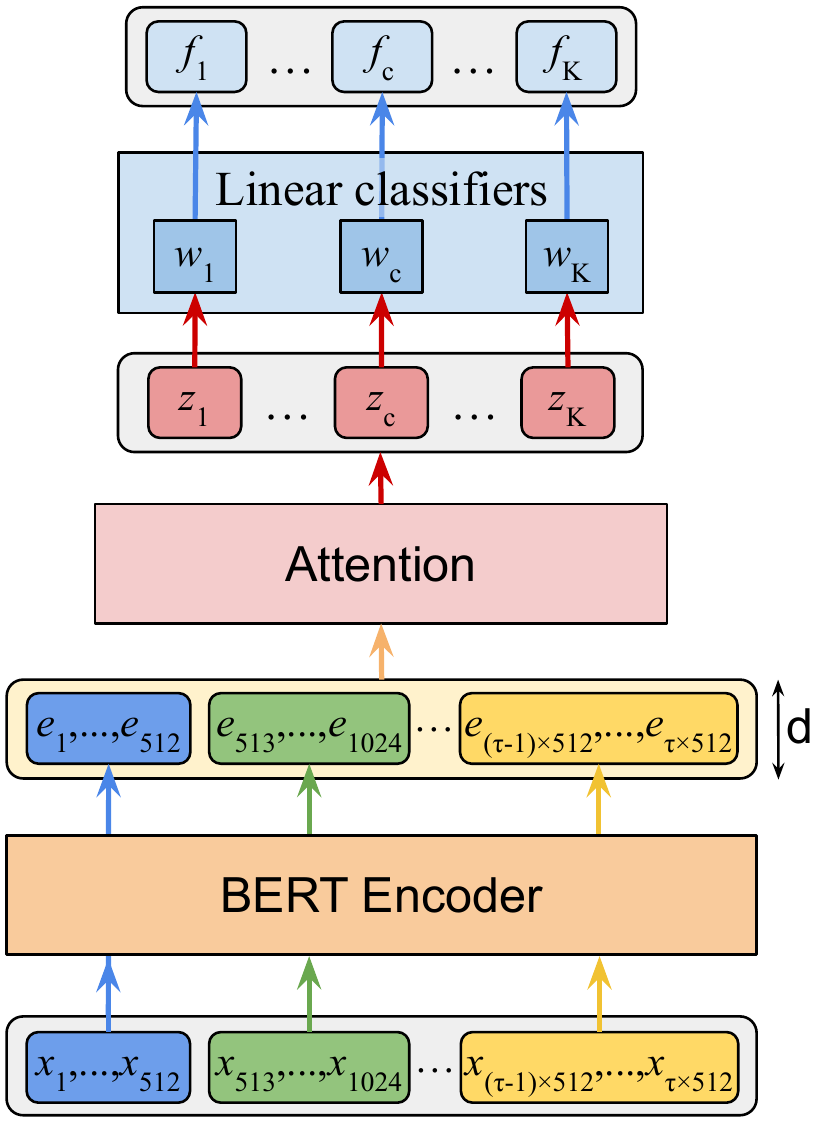}
    \caption{Model architecture proposed for handling long text inputs.}
    \label{fig:model}
\end{figure}

\section{Method}
\label{sec:method}

In this section we explain our method for building a model to predict medical codes. As illustrated in Figure~\ref{fig:model}, our model consists of an encoder that calculates token-level representation of the input text. This can be done in various ways, e.g. \citet{mullenbach2018explainable} used word2vec and a CNN layer to calculate word-level representations. We choose the BERT language model for this purpose. A class-specific representation of the document is then calculated using class-specific attention vectors, similar to \citet{mullenbach2018explainable}. For $d$-dimensional token representations and $K$ classes, this layer requires $d \times K$ parameters. Linear binary classifiers are built on top of the document representation to produce the probability that the document belongs to any of the $K$ classes. This layer requires $(d+1) \times K$ parameters (one scalar for the offset).

Let $X = [x_1, \ldots, x_s]$ denote the tokenized input sequence with $s$ tokens. Let $e_1(X), \ldots, e_s(X)$ denote the representation of tokens $1, \ldots, s$ obtained from an encoder. That is,
\beqann
  e_i(X) = \phi(x_i | X), \hspace{5mm} i \in \{1, \ldots, s\},  
\eeqann
where $\phi$ is an encoder, such as BERT, that returns a context-dependent representation for each token. For each class $c$, token-level representations are combined into a single vector that represents the entire document using the attention mechanism:
\beqann
  z_c(X) = \sum_{i=1}^s \alpha_{c,i}(X) e_i(X),
\eeqann
where
\beqa
  &&\alpha_{c,i}(X) = \frac{\exp\left( \langle e_i(X), q_c \rangle \right)}{ \sum_{j=1}^s \exp\left( \langle e_j(X), q_c \rangle \right) },\\
  &&i \in \{1, \ldots, s\}, \nonumber
  \label{eq:alpha}
\eeqa
are the normalized attention coefficients and $\inl \cdot, \cdot \inr$ denotes inner product, and $q_c$ is the $d$-dimensional attention vector for class $c$.
The predicted probability of the model for class $c$ is calculated by
\beqann
    f_c(X) = \sigma \left( \inl z_c(X), w_c \inr + b_c \right),
\eeqann
where $w_c$ is the weight vector for class $c$, $b_c$ is the scalar offset for class $c$, and $\sigma$ is the sigmoid function. 

\subsection{Handling long text}
\label{sec:longtext}

Language models such as BERT can handle input text up to a certain length. For example, BERT can take input of at most 512 tokens. While it is possible to pre-train the model on longer sequences (mostly to learn useful positional embedding vectors), memory requirement grows quadratically with input size. So pre-training a BERT model on longer text is not scalable.

There are transformer-based models that can handle long sequences, such as BigBird \citep{zaheer2020big}, ETC \citep{ainslie2020etc}, Longformer \citep{beltagy2020longformer}, and LongT5 \citep{guo2021longt5}. There are a few factors that limit their usability in the medical coding task. For example, these models are usually designed to train on TPU, so training on GPU is often a slow process, if feasible, especially for longer sequences. Also, pre-trained checkpoints of these models are limited, unlike the BERT models that have many pre-trained variants including those pre-trained on medical text.

In this paper, we propose a simple idea, which enables us to use a vanilla BERT model on long sequences. Inspired by the local attention feature of CNN models, we propose to split the input text into (optionally overlapping) segments of 512 tokens. These segments are passed sequentially to a BERT model, and the token representations are concatenated to form $[e_1(X), ..., e_{512}(X), e_{513}(X), ..., e_{1024}(X), ...]$. One may argue that a limitation of this approach is that the token representations are calculated with a 512-token attention span. However, we have observed that in practice this method performs well. In fact, we conjecture that in many cases short snippets of text (as evidence) are sufficient for assigning the correct ICD codes to the input document. Algorithm~\ref{alg:longbert} shows the training procedure.

\begin{algorithm*}[t]
    \caption{Training on a single example.}
    \begin{algorithmic}[1]
        \State \textbf{Input:} tokenized input text of length $s$: $X = [x_1, \ldots, x_s]$, sparse binary label vector $Y = [y_1, \ldots, y_K]$ for $K$ classes, where $y_c = 1$ if the example belongs to class $c$, and $0$ otherwise.
        \State Pad input text $X = [x_1, \ldots, x_s]$ to length $\tau \times 512$ to obtain $X' = [x_1, \ldots, x_s, \ldots, x_{\tau \times 512}]$, where $\tau = \lceil s/512 \rceil$.
        \State Split $X'$ into segments of 512 tokens: $S_1 = [x_1, \ldots, x_{512}], S_2 = [x_{513}, \ldots, x_{1024}], \ldots, S_{\tau}$.
        \State Pass $S_i\text{'s}, \hspace{1mm} i \in \{1, \ldots, \tau\}$ sequentially to the BERT module and obtain the corresponding token representations.
        \State Concatenate token representations from all sequences to obtain $[e_1, \ldots, e_s, \ldots, e_{\tau \times 512}]$.
        \State Calculate class-specific document representations by $z_c(X) = \sum_{i=1}^s \alpha_{c,i}(X) e_i(X)$, with $\alpha_{c,i}$ from Eq.~\ref{eq:alpha}.
        \State Calculate model predictions for all classes: $f_c(X) = \sigma \left( \inl z_c(X), w_c \inr + b_c \right), \hspace{1mm} c \in {1, \ldots, K}$.
        \State Calculate and apply gradient updates for loss function $\sum_{c=1}^K \loss(y_c, f_c(X))$, where $\loss$ is binary cross-entropy.
    \end{algorithmic}
    \label{alg:longbert}
\end{algorithm*}

\section{Evaluation}

We evaluate the accuracy of the proposed method with several sequence lengths and compare it against the CAML method \citep{mullenbach2018explainable}, which is one of the prominent CNN-based methods for ICD coding.

\subsection{Data sets}

For this task we chose the publicly-available MIMIC-III \citep{johnson2016mimic} and MIMIC-IV \citep{johnson2020mimic} data sets. MIMIC-III is a large de-identified data set of over 40,000 patients admitted to intensive care units at the Beth Israel Deaconess Medical Center. The data set contains structured and unstructured data, including lab measurements, vital signs, medications, clinical notes, etc. Following previous studies, we focus on predicting ICD codes for discharge summaries where each note corresponds to a hospital stay event. MIMIC-IV is an update to MIMIC-III, which incorporates contemporary data. It is sourced from two in-hospital database systems: a custom hospital wide EHR and an ICU specific clinical information system.

Each discharge summary in MIMIC-III is manually coded by human coders with one or more ICD-9 codes that specify diagnoses and procedures of that particular stay.
The data set contains 8,921 unique ICD-9 codes, including 6,918 diagnosis and 2,003 procedure codes. There are patients with
multiple admissions and therefore multiple discharge summaries. To be consistent with the previous studies and to ensure that all of the notes of a patient are assigned to one of train/validation/test sets we use the data split provided by \citet{mullenbach2018explainable}. This results in 47,724 discharge summaries for training, 1,632 summaries and 3,372 summaries for validation and test sets respectively.

The discharge summaries in MIMIC-IV are additionally labeled with ICD-10 codes. At the time of writing this paper the MIMIC-Note module, which contains the discharge summaries, is not yet publicly available. In our experiments we only consider the ICD-10 diagnosis set, which contains 72,748 codes in the data set.

\begin{figure*}
     \centering
     \begin{subfigure}
        \centering
        \includegraphics[width=0.49\textwidth]{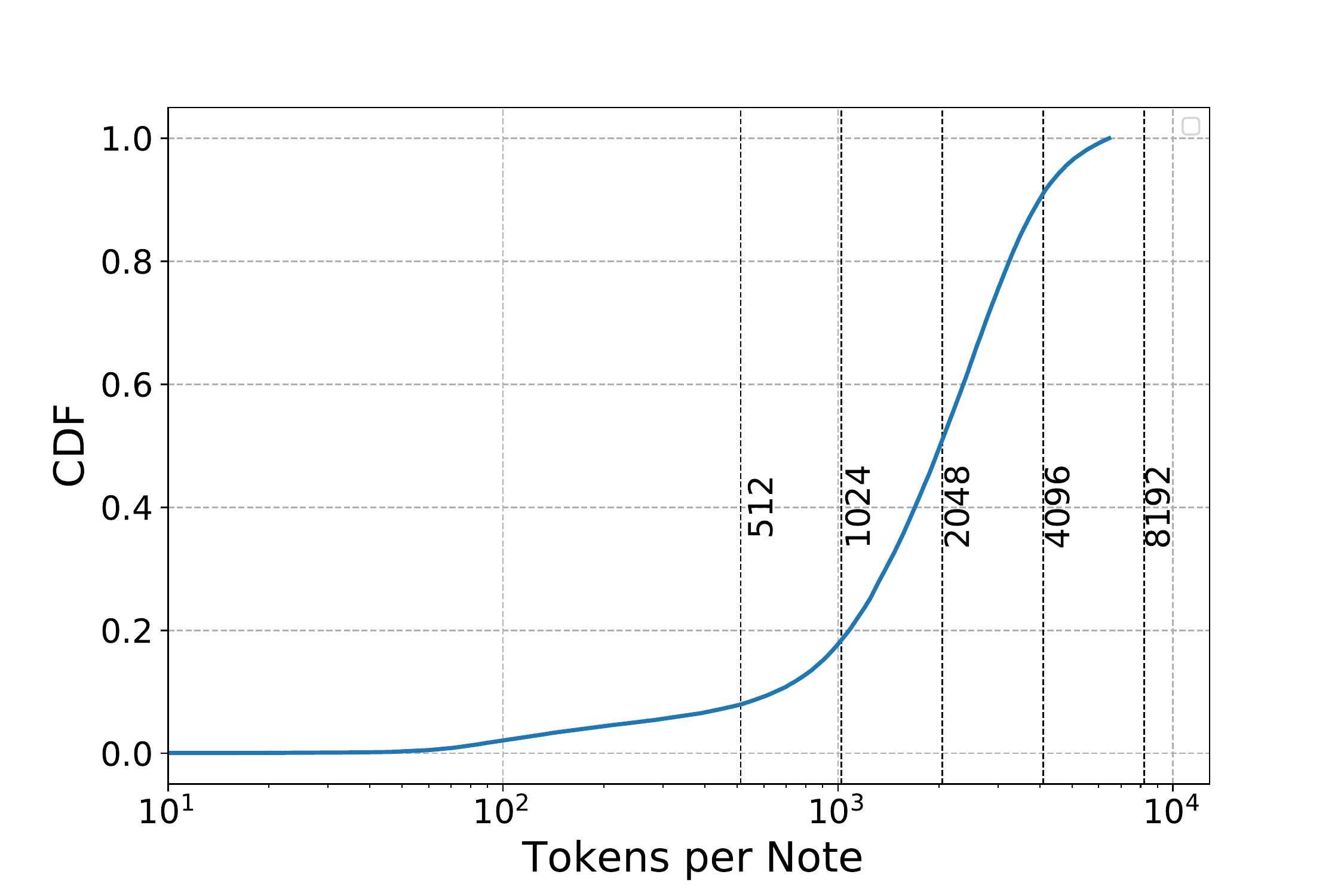}
     \end{subfigure}
     \hfill
     \begin{subfigure}
        \centering
        \includegraphics[width=0.49\textwidth]{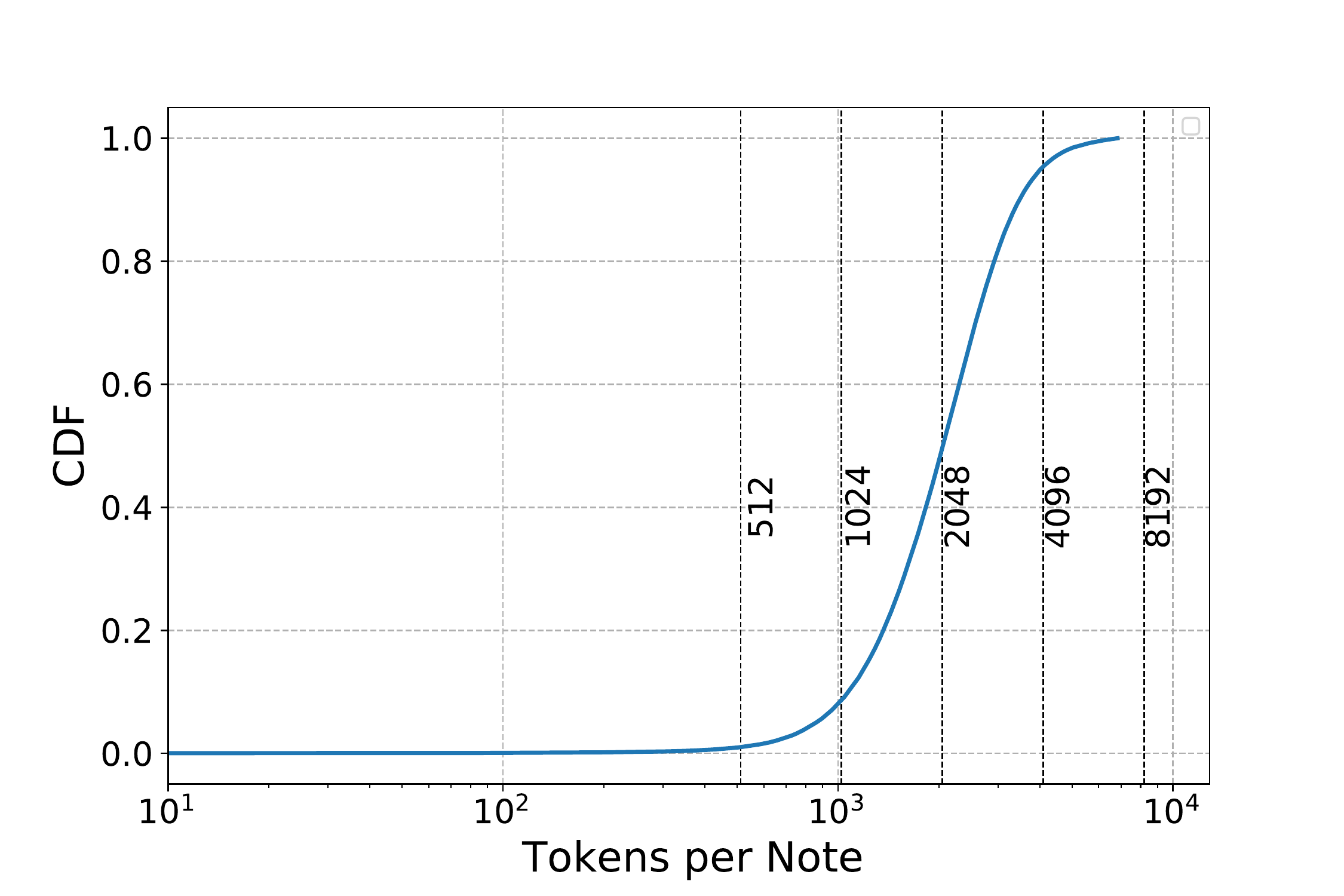}
     \end{subfigure}
     \caption{Cumulative distribution function (CDF) of the number of tokens per note for MIMIC-III (left) and MIMIC-IV (right) data sets.}
     \label{fig:cdf}
\end{figure*}

For tokenizing text we used the standard BERT vocabulary and tokenizer \citep{devlin2018bert}. Figure~\ref{fig:cdf} shows the cumulative distribution function of the number of tokens per note for MIMIC-III and MIMIC-IV.

\subsection{Models}

Our classification model uses a BERT language model with the method described in Section~\ref{sec:method}. We dub this model \emph{LongBERT} below. The BERT checkpoint we use in the experiments is a model with 2 transformer blocks and 256-dimensional embedding vectors. The checkpoint can be downloaded from TensorFlow Hub.\footnote{\url{https://tfhub.dev/tensorflow/small_bert/bert_en_uncased_L-2_H-256_A-4/2}} 

The baseline model (\emph{BERT-baseline}) was trained and evaluated on the first 512 tokens of input text. To measure the impact of sequence length we trained and evaluated similar models on the first $s$ tokens of each note, with $s \in \{1024, 2048, 4096, 8192\}$. All BERT parameters and the additional attention and classification parameters were fine-tuned during training. We used the Adam optimizer \citep{kingma2014adam} with a learning rate of 2e-4. The batch size was set to 4 in all experiments, except for the models trained with the sequence length of 8192 which were trained with the batch size of 2 to avoid running out of memory. The models were trained for 1 million steps (each step is one batch). No hyper-parameter tuning was performed except for the number of training steps. The best model corresponds to the training step that achieves the highest validation micro F1 score.

We compare these models against a CAML model trained on sequences of 2500 words following \citet{mullenbach2018explainable}. The hyper-parameters were set according to the optimal values obtained in \citet{mullenbach2018explainable}. Training was performed for 1 million steps, and the best model was selected according to validation micro F1 score.

Following previous work, in the MIMIC-III experiments, training and evaluation was performed on the full ICD-9 label set as well as the 50 most frequent codes. In the MIMIC-IV experiment, we consider only the ICD-10 diagnosis codes. Each ICD code has its own attention and classification weight vectors in the models. Table~\ref{tab:num-parameters} breaks down the number of parameters of the models in the experiments.

\begin{table*}
\centering
\begin{tabular}{l|c|c|c}
& \textbf{MIMIC-III full} & \textbf{MIMIC-III top 50} & \textbf{MIMIC-IV diagnosis} \\
\toprule
{Language model} & 9,591,040 & 9,591,040 & 9,591,040 \\ 
{Attention layer} & 2,283,776 & 12,800 & 18,623,488 \\ 
{Classification layer} & 2,292,697 & 12,850 & 18,696,236 \\
\midrule
{Total} & 14,167,513 & 9,616,690 & 46,910,764 \\ 
\bottomrule
\end{tabular}
\caption{Breakdown of the number of parameters of BERT-baseline and LongBERT with 2 transformer blocks and 256-dimensional embedding vectors. MIMIC-III full contains 8,921 classes, and MIMIC-IV diagnosis contains 72,748 classes.}
\label{tab:num-parameters}
\end{table*}

\subsection{Evaluation metrics}

Our primary evaluation metric is micro-averaged F1 (micro F1 for short). Micro-averaged values are calculated by treating each code as a (binary) label for each note. That is, each (note, code) pair is counted as one instance for calculating the metrics. Let,
\beqann
\text{micro precision} &=& \frac{\sum_{x,c} TP(x, c)}{\sum_{x,c} TP(x, c) + FP(x, c)},\\
\text{micro recall} &=& \frac{\sum_{x,c} TP(x, c)}{\sum_{x,c} TP(x, c) + FN(x, c)},
\eeqann
where $TP(x, c) = 1$ if class $c$ is a true positive prediction for note $x$ and 0 otherwise. $FP(x, c)$ (false positive) and $FN(x, c)$ (false negative) are defined analogously. Finally, micro F1 is the harmonic mean of micro precision and micro recall:
\beqann
\text{micro F1} &=& 2 \hspace{2mm} \frac{\text{micro precision} \times \text{micro recall}}{\text{micro precision} + \text{micro recall}}.
\eeqann
The optimal threshold on model predictions, which is used to calculate $TP/FP/FN$ counts, is obtained by a grid search to maximize the validation set F1 score.

Additionally we report precision-recall AUC (PR-AUC), and ROC-AUC. In contrast to F1 score, these metrics are independent of a specific operating point and provide an aggregated view of model accuracy.

\subsection{Results}

Table~\ref{tab:mimic3-full-code} shows the results of the LongBERT and CAML models on the MIMIC-III full-code test set. Table~\ref{tab:mimic4-diagnosis} shows accuracy metrics obtained on the MIMIC-IV diagnosis code data set. Bold numbers represent the best value of each metric. A clear trend observed in both data sets is that as the sequence length of LongBERT increases, the accuracy of the model improves. These results demonstrate that the capability to process long text is critical in achieving high accuracy.

The LongBERT models with sequence lengths of 4096 and 8192 both outperform the CAML model. This finding contradicts the previous finding of \citet{ji2021does}. While their hierarchical attention proposal and our method both handle long text by breaking it into segments of 512 tokens, one key difference is that they use the CLS token representation from each segment, whereas we use individual token representations. The best MIMIC-III full-code performance reported in \citet{ji2021does} was F1 = 0.47 with BioBERT full-text \citep{lee2020biobert} checkpoint and hierarchical attention, while our small vanilla BERT model with sequence length of 8192 achieves F1 = 0.5680. These results show that with a proper modeling approach transformer-based models are indeed capable of outperforming CNN-based models in ICD coding.

\begin{table*}
\centering
\begin{tabular}{p{22mm}|c|c|c|c|c|c}
\toprule
 & \textbf{Seq. length} & \textbf{Micro F1} & \textbf{Precision} & \textbf{Recall} & \textbf{PR-AUC} & \textbf{ROC-AUC} \\
\midrule
CAML & 2500 (words) & 0.5465 & 0.5973 & 0.5036 & 0.5361 & \textbf{0.9831} \\
BioBERT full-text \citep{ji2021does} & entire note & 0.470 & N/A & N/A & N/A & 0.974 \\ 
\midrule
BERT-baseline & 512 & 0.4149 & 0.4769 & 0.3672 & 0.3793 & 0.9745 \\
LongBERT & 1024 & 0.4697 & 0.5421 & 0.4144 & 0.4309 & 0.9766 \\
LongBERT & 2048 & 0.5036 & 0.5777 & 0.4463 & 0.4703 & 0.9794 \\
LongBERT & 4096 & 0.5514 & 0.6038 & 0.5074 & 0.5305 & 0.9820 \\
LongBERT & 8192 & \textbf{0.5680} & \textbf{0.6148} & \textbf{0.5278} & \textbf{0.5402} & 0.9827 \\
\bottomrule
\end{tabular}
\caption{Accuracy metrics in the MIMIC-III full-code experiment.}
\label{tab:mimic3-full-code}
\end{table*}

\begin{table*}
\centering
\begin{tabular}{l|c|c|c|c|c|c}
\toprule
 & \textbf{Seq. length} & \textbf{Micro F1} & \textbf{Precision} & \textbf{Recall} & \textbf{PR-AUC} & \textbf{ROC-AUC} \\
\midrule
CAML & 2500 (words) & 0.5439 & 0.5739 & 0.5169 & 0.5313 & \textbf{0.9889} \\
\midrule
BERT-baseline & 512 & 0.4010 & 0.4298 & 0.3757 & 0.3580 & 0.9883 \\
LongBERT & 1024 & 0.4607 & 0.5094 & 0.4205 & 0.4254 & 0.9839 \\
LongBERT & 2048 & 0.4852 & 0.5268 & 0.4497 & 0.4559 & 0.9852 \\
LongBERT & 4096 & 0.5635 & 0.5925 & 0.5371 & 0.5450 & 0.9850 \\
LongBERT & 8192 & \textbf{0.5703} & \textbf{0.6046} & \textbf{0.5397} & \textbf{0.5517} & 0.9871 \\
\bottomrule
\end{tabular}
\caption{Accuracy metrics in the MIMIC-IV diagnosis experiment.}
\label{tab:mimic4-diagnosis}
\end{table*}

\paragraph{MIMIC-III top 50.}
Following previous work, we also trained and evaluated the models on the MIMIC-III 50 most frequent codes. Table~\ref{tab:mimic3-top50-code} shows the results. Similar to the full-code case we observe that processing longer segments results in higher accuracy.

In this case, however, there is no clear winner between LongBERT and CAML. While LongBERT achieves a higher micro F1 score, the CAML model has a higher PR-AUC.
We conjecture that the smaller performance difference between the two models in this experiment compared to the full-code experiment is due to the amount of information in the data sets. By removing many of the labels in the top-50 experiment we essentially remove information. This information is more helpful to larger models (i.e. transformers) than smaller models, such as CAML. As a result, we observe a larger performance gap in the full-code experiment between LongBERT and CAML.

We also note that the accuracy numbers of the CAML model in this experiment are higher than those reported in \citet{mullenbach2018explainable}. One difference here is that we do not discard notes that aren't assigned any of the top 50 codes as was done in the original paper. Such notes are used as negative examples for the top 50 codes. Therefore our data set contains more negative examples than the data set used in \citet{mullenbach2018explainable}.

\begin{table*}[t!]
\centering
\begin{tabular}{l|c|c|c|c|c|c}
\toprule
 & \textbf{Seq. length} & \textbf{Micro F1} & \textbf{Precision} & \textbf{Recall} & \textbf{PR-AUC} & \textbf{ROC-AUC} \\
\midrule
CAML & 2500 (words) & 0.6390 & \textbf{0.6506} & 0.6278 & \textbf{0.6410} & 0.9102 \\
\midrule
BERT-baseline & 512 & 0.5027 & 0.5367 & 0.4727 & 0.5117 & 0.8360 \\
LongBERT & 1024 & 0.5568 & 0.5923 & 0.5252 & 0.5406 & 0.8560 \\
LongBERT & 2048 & 0.5908 & 0.5987 & 0.5832 & 0.5604 & 0.8834 \\
LongBERT & 4096 & 0.6375 & 0.6157 & 0.6609 & 0.6229 & 0.9115 \\
LongBERT & 8192 & \textbf{0.6522} & 0.6417 & \textbf{0.6629} & 0.6303 & \textbf{0.9181} \\
\bottomrule
\end{tabular}
\caption{Accuracy metrics in the MIMIC-III top-50 experiment.}
\label{tab:mimic3-top50-code}
\end{table*}

\section{Discussion}

Most of the existing BERT models pre-trained on generic or medical text can take input segments of up to 512 tokens. Clinical notes, however, are much longer than this limit. To deal with this limitation, much of the existing works in automated ICD coding that use BERT limit the input to the model by truncating the text or selecting specific spans of text. This results in loss of information and poor performance.

In this paper we proposed a simple method to apply BERT models to sequences longer than 512 tokens. Our method is simple and consists of two key components: (i) apply BERT sequentially to (optionally overlapping) segments of 512 tokens, and (ii) concatenate token-level representations from all segments, and combine them using a class-specific attention layer.

We demonstrated that processing long text sequences minimizes information loss and is critical for achieving high performance in automated ICD coding. We also showed that contrary to previous findings, this method with even a small vanilla BERT model outperforms CNN-based methods, and achieves competitive performance.

Future steps include evaluating medical variants of BERT, and exploring other transformer-based architectures that were designed to handle long sequences.

\section*{Limitations}
While our method enables the processing of text longer than 512 tokens, one of the limitations of this approach is that context-dependent token representations are still calculated using a window of 512 tokens. Despite good performance of this method in practice, there could be cases where a context window of longer than 512 must be used to make accurate predictions.

Furthermore, while our method reduces computational complexity from quadratic (in sequence length) to linear, the memory requirement of the model could still be prohibitive in certain cases. For instance, for sequence length of 8192, and a small BERT checkpoint with only two transformer blocks we had to reduce batch size to 2 in order to train the models. Using a larger BERT checkpoint for long sequences requires more memory and multiple GPUs, which increases the cost of compute.

\bibliography{references}

\begin{thebibliography}{25}
\expandafter\ifx\csname natexlab\endcsname\relax\def\natexlab#1{#1}\fi

\bibitem[{Ainslie et~al.(2020)Ainslie, Ontanon, Alberti, Cvicek, Fisher, Pham,
  Ravula, Sanghai, Wang, and Yang}]{ainslie2020etc}
Joshua Ainslie, Santiago Ontanon, Chris Alberti, Vaclav Cvicek, Zachary Fisher,
  Philip Pham, Anirudh Ravula, Sumit Sanghai, Qifan Wang, and Li~Yang. 2020.
\newblock Etc: Encoding long and structured inputs in transformers.
\newblock \emph{arXiv preprint arXiv:2004.08483}.

\bibitem[{Amin et~al.(2019)Amin, Neumann, Dunfield, Vechkaeva, Chapman, and
  Wixted}]{amin2019mlt}
Saadullah Amin, G{\"u}nter Neumann, Katherine Dunfield, Anna Vechkaeva,
  Kathryn~Annette Chapman, and Morgan~Kelly Wixted. 2019.
\newblock Mlt-dfki at clef ehealth 2019: Multi-label classification of icd-10
  codes with bert.
\newblock In \emph{CLEF (Working Notes)}, pages 1--15.

\bibitem[{Baumel et~al.(2018)Baumel, Nassour-Kassis, Cohen, Elhadad, and
  Elhadad}]{baumel2018multi}
Tal Baumel, Jumana Nassour-Kassis, Raphael Cohen, Michael Elhadad, and
  No{\'e}mie Elhadad. 2018.
\newblock Multi-label classification of patient notes: case study on icd code
  assignment.
\newblock In \emph{Workshops at the thirty-second AAAI conference on artificial
  intelligence}.

\bibitem[{Beltagy et~al.(2020)Beltagy, Peters, and
  Cohan}]{beltagy2020longformer}
Iz~Beltagy, Matthew~E Peters, and Arman Cohan. 2020.
\newblock Longformer: The long-document transformer.
\newblock \emph{arXiv preprint arXiv:2004.05150}.

\bibitem[{Biseda et~al.(2020)Biseda, Desai, Lin, and
  Philip}]{biseda2020prediction}
Brent Biseda, Gaurav Desai, Haifeng Lin, and Anish Philip. 2020.
\newblock Prediction of icd codes with clinical bert embeddings and text
  augmentation with label balancing using mimic-iii.
\newblock \emph{arXiv preprint arXiv:2008.10492}.

\bibitem[{Cao et~al.(2020)Cao, Chen, Liu, Zhao, Liu, and
  Chong}]{cao2020hypercore}
Pengfei Cao, Yubo Chen, Kang Liu, Jun Zhao, Shengping Liu, and Weifeng Chong.
  2020.
\newblock Hypercore: Hyperbolic and co-graph representation for automatic icd
  coding.
\newblock In \emph{Proceedings of the 58th Annual Meeting of the Association
  for Computational Linguistics}, pages 3105--3114.

\bibitem[{Chen(2015)}]{chen2015convolutional}
Yahui Chen. 2015.
\newblock Convolutional neural network for sentence classification.
\newblock Master's thesis, University of Waterloo.

\bibitem[{Devlin et~al.(2018)Devlin, Chang, Lee, and
  Toutanova}]{devlin2018bert}
Jacob Devlin, Ming-Wei Chang, Kenton Lee, and Kristina Toutanova. 2018.
\newblock Bert: Pre-training of deep bidirectional transformers for language
  understanding.
\newblock \emph{arXiv preprint arXiv:1810.04805}.

\bibitem[{Guo et~al.(2021)Guo, Ainslie, Uthus, Ontanon, Ni, Sung, and
  Yang}]{guo2021longt5}
Mandy Guo, Joshua Ainslie, David Uthus, Santiago Ontanon, Jianmo Ni, Yun-Hsuan
  Sung, and Yinfei Yang. 2021.
\newblock Longt5: Efficient text-to-text transformer for long sequences.
\newblock \emph{arXiv preprint arXiv:2112.07916}.

\bibitem[{Ji et~al.(2021)Ji, H{\"o}ltt{\"a}, and Marttinen}]{ji2021does}
Shaoxiong Ji, Matti H{\"o}ltt{\"a}, and Pekka Marttinen. 2021.
\newblock Does the magic of bert apply to medical code assignment? a
  quantitative study.
\newblock \emph{Computers in Biology and Medicine}, 139:104998.

\bibitem[{Johnson et~al.(2020)Johnson, Bulgarelli, Pollard, Horng, Celi, and
  Mark}]{johnson2020mimic}
Alistair Johnson, Lucas Bulgarelli, Tom Pollard, Steven Horng, Leo~Anthony
  Celi, and Roger Mark. 2020.
\newblock Mimic-iv.
\newblock \emph{version 0.4). PhysioNet. https://doi. org/10.13026/a3wn-hq05}.

\bibitem[{Johnson et~al.(2016)Johnson, Pollard, Shen, Lehman, Feng, Ghassemi,
  Moody, Szolovits, Anthony~Celi, and Mark}]{johnson2016mimic}
Alistair~EW Johnson, Tom~J Pollard, Lu~Shen, Li-wei~H Lehman, Mengling Feng,
  Mohammad Ghassemi, Benjamin Moody, Peter Szolovits, Leo Anthony~Celi, and
  Roger~G Mark. 2016.
\newblock Mimic-iii, a freely accessible critical care database.
\newblock \emph{Scientific data}, 3(1):1--9.

\bibitem[{Kingma and Ba(2014)}]{kingma2014adam}
Diederik~P Kingma and Jimmy Ba. 2014.
\newblock Adam: A method for stochastic optimization.
\newblock \emph{arXiv preprint arXiv:1412.6980}.

\bibitem[{Lee et~al.(2020)Lee, Yoon, Kim, Kim, Kim, So, and
  Kang}]{lee2020biobert}
Jinhyuk Lee, Wonjin Yoon, Sungdong Kim, Donghyeon Kim, Sunkyu Kim, Chan~Ho So,
  and Jaewoo Kang. 2020.
\newblock Biobert: a pre-trained biomedical language representation model for
  biomedical text mining.
\newblock \emph{Bioinformatics}, 36(4):1234--1240.

\bibitem[{Li and Yu(2020)}]{li2020icd}
Fei Li and Hong Yu. 2020.
\newblock Icd coding from clinical text using multi-filter residual
  convolutional neural network.
\newblock In \emph{Proceedings of the AAAI Conference on Artificial
  Intelligence}, volume 34(05), pages 8180--8187.

\bibitem[{Medori and Fairon(2010)}]{medori2010machine}
Julia Medori and C{\'e}drick Fairon. 2010.
\newblock Machine learning and features selection for semi-automatic icd-9-cm
  encoding.
\newblock In \emph{Proceedings of the NAACL HLT 2010 Second Louhi Workshop on
  Text and Data Mining of Health Documents}, pages 84--89.

\bibitem[{Mullenbach et~al.(2018)Mullenbach, Wiegreffe, Duke, Sun, and
  Eisenstein}]{mullenbach2018explainable}
James Mullenbach, Sarah Wiegreffe, Jon Duke, Jimeng Sun, and Jacob Eisenstein.
  2018.
\newblock Explainable prediction of medical codes from clinical text.
\newblock \emph{arXiv preprint arXiv:1802.05695}.

\bibitem[{Pascual et~al.(2021)Pascual, Luck, and
  Wattenhofer}]{pascual2021towards}
Damian Pascual, Sandro Luck, and Roger Wattenhofer. 2021.
\newblock Towards bert-based automatic icd coding: Limitations and
  opportunities.
\newblock \emph{arXiv preprint arXiv:2104.06709}.

\bibitem[{Perotte et~al.(2014)Perotte, Pivovarov, Natarajan, Weiskopf, Wood,
  and Elhadad}]{perotte2014diagnosis}
Adler Perotte, Rimma Pivovarov, Karthik Natarajan, Nicole Weiskopf, Frank Wood,
  and No{\'e}mie Elhadad. 2014.
\newblock Diagnosis code assignment: models and evaluation metrics.
\newblock \emph{Journal of the American Medical Informatics Association},
  21(2):231--237.

\bibitem[{Shi et~al.(2017)Shi, Xie, Hu, Zhang, and Xing}]{shi2017towards}
Haoran Shi, Pengtao Xie, Zhiting Hu, Ming Zhang, and Eric~P Xing. 2017.
\newblock Towards automated icd coding using deep learning.
\newblock \emph{arXiv preprint arXiv:1711.04075}.

\bibitem[{Singh et~al.(2020)Singh, Guntu, Bhimireddy, Gichoya, and
  Purkayastha}]{singh2020multi}
AK~Singh, Mounika Guntu, Ananth~Reddy Bhimireddy, Judy~W Gichoya, and Saptarshi
  Purkayastha. 2020.
\newblock Multi-label natural language processing to identify diagnosis and
  procedure codes from mimic-iii inpatient notes.
\newblock \emph{arXiv preprint arXiv:2003.07507}.

\bibitem[{Tay et~al.(2020)Tay, Dehghani, Bahri, and Metzler}]{tay2020efficient}
Yi~Tay, Mostafa Dehghani, Dara Bahri, and Donald Metzler. 2020.
\newblock Efficient transformers: A survey.
\newblock \emph{ACM Computing Surveys (CSUR)}.

\bibitem[{Vaswani et~al.(2017)Vaswani, Shazeer, Parmar, Uszkoreit, Jones,
  Gomez, Kaiser, and Polosukhin}]{vaswani2017attention}
Ashish Vaswani, Noam Shazeer, Niki Parmar, Jakob Uszkoreit, Llion Jones,
  Aidan~N Gomez, {L}ukasz Kaiser, and Illia Polosukhin. 2017.
\newblock Attention is all you need.
\newblock \emph{Advances in neural information processing systems}, 30.

\bibitem[{Zaheer et~al.(2020)Zaheer, Guruganesh, Dubey, Ainslie, Alberti,
  Ontanon, Pham, Ravula, Wang, Yang et~al.}]{zaheer2020big}
Manzil Zaheer, Guru Guruganesh, Kumar~Avinava Dubey, Joshua Ainslie, Chris
  Alberti, Santiago Ontanon, Philip Pham, Anirudh Ravula, Qifan Wang, Li~Yang,
  et~al. 2020.
\newblock Big bird: Transformers for longer sequences.
\newblock \emph{Advances in Neural Information Processing Systems},
  33:17283--17297.

\bibitem[{Zhang et~al.(2020)Zhang, Liu, and Razavian}]{zhang2020bert}
Zachariah Zhang, Jingshu Liu, and Narges Razavian. 2020.
\newblock Bert-xml: Large scale automated icd coding using bert pretraining.
\newblock \emph{arXiv preprint arXiv:2006.03685}.

\end{thebibliography}
\bibliographystyle{acl_natbib}

\end{document}